\PassOptionsToPackage{table}{xcolor}
\documentclass[sigconf]{acmart}

\usepackage{amsmath,amsfonts,bm}

\def\eqref#1{equation~\ref{#1}}

\def\1{\bm{1}}

\DeclareMathAlphabet{\mathsfit}{\encodingdefault}{\sfdefault}{m}{sl}
\SetMathAlphabet{\mathsfit}{bold}{\encodingdefault}{\sfdefault}{bx}{n}

\usepackage{booktabs}
\usepackage{graphicx}
\usepackage{multirow}
\usepackage{subcaption}
\newcommand{\OrgCall}[1]{\mathcal{C}_{M}\!\left(#1\right)}
\newcommand{\OrgExec}[2]{\mathcal{X}\!\left(#1,#2\right)}
\newcommand{\OrgState}[2]{\mathcal{S}\!\left(#1,#2\right)}
\newcommand{\OrgFit}[1]{\mathcal{F}_{#1}}
\hypersetup{hidelinks}
\setcopyright{none}
\acmDOI{}
\acmISBN{}
\acmConference[WWW '27]{The ACM Web Conference 2027}{May 10--14, 2027}{Dublin, Ireland}
\acmYear{2027}
\copyrightyear{2027}

\definecolor{opendgobaseline}{HTML}{FCEDE5}
\definecolor{opendgofixed}{HTML}{EAF2FB}
\definecolor{opendgolearned}{HTML}{E5F6F4}
\definecolor{opendgopanel}{HTML}{F2F2F2}

\definecolor{opendgobest}{HTML}{D9D9D9}
\definecolor{opendgosecond}{HTML}{EEEEEE}

\title{OpenDGO: Benchmarking Multi-Agent Organizations Across Diverse Application Domains}

\author{Xunkai Li*, Miao Zhang*, Zekai Chen*, Xuaner Chen, Yu liu, Rui Qian, Xun Wu, Hongchao Qin, Rong-Hua Li}

\begin{document}

\begin{abstract}
Large language model (LLM)-based multi-agent systems are increasingly applied across diverse domains, yet different applications require different forms of agent organization. Existing studies typically adopt predefined organizations, such as Star, Chain, and Tree, or learn communication graphs to improve performance on specific tasks. However, these methods are largely designed and evaluated within a limited set of application domains, providing little understanding of whether an organization effective in one domain remains suitable for others with different collaboration requirements. We present \textbf{OpenDGO}, a controlled benchmark and analysis framework covering six real-world domains and three collaboration settings: Sequential Dependency, Parallel Decomposition, and Information Integration. Under matched tasks, agents, tools, and resource limits, OpenDGO compares fixed and learned organizations while separately examining communication graphs, role assignments, input assignments, resource budgets, and component failures. Our results show that there is no ``one-organization-for-all'' solution: preferred organizations vary across applications, and even the same communication graph can yield different outcomes under different role and input assignments. These findings highlight the need for application-dependent organization design, aiming to understand \emph{which organization is suitable for a given application, under what conditions, and why}.
\end{abstract}


\maketitle


\section{Introduction}

Large language model (LLM)-based multi-agent systems (MASs) are increasingly used in web-centric applications, such as web search, online recommendation, and document question answering, by coordinating multiple specialized agents to solve complex problems~\cite{zhu2025multiagentbench,yang2026magic,yang2026mace}. A basic design choice is how these agents should be organized, including which agents can communicate with each other and which role each agent plays in the communication graph (e.g., which agent is responsible for a specific function, such as implementation, testing, or verification). Common choices include centralized Star structures, ordered Chain structures, hierarchical Trees, and irregular graphs learned automatically for a given task. Different application domains naturally favor different forms of organization. In Coding, issue analysis, implementation, and testing form an ordered workflow, making Chain- or Tree-like organizations intuitive choices for passing intermediate results between stages. Finance similarly requires extracted numbers to move reliably through calculation and verification. Legal and Scientific Research instead require several agents to examine different documents, rules, or claims before their conclusions are combined, which can benefit from more distributed communication. Recommendation often allows user history, reviews, and item attributes to be analyzed in parallel before aggregation, making highly centralized sequential communication less necessary. These application domains therefore place different demands on the organization of an agent team, making a single organization unlikely to work equally well across all of them. The central question is thus \emph{which type of multi-agent organization is appropriate for a given application domain, and why}.

Existing studies have mainly focused on evaluating which organizations perform well within a specific application domain, while paying little attention to whether the same organizations remain effective in other domains with different collaboration requirements~\cite{zhang2025gdesigner,li2026argdesigner,chen2026goagent}. For general-purpose MASs, this distinction matters: an organization suited to the sequential workflow of Coding may not work equally well for Legal analysis or Recommendation, where agents need to collaborate differently. Furthermore, the communication graph is not the only factor affecting performance. Even within a fixed Tree, placing a verifier at the center rather than at a peripheral node changes how it communicates with other agents. Similarly, assigning different inputs, such as code files or financial tables, to different agents can affect their collaboration. To examine these factors, we evaluate the same organizations across diverse domains, keeping the source task, agent model, tools, and resource limits fixed within each comparison while separately varying the communication graph, role assignment, and input assignment. This allows us to assess how organization preferences change across domains and identify the factors behind these differences.

Our motivating observations in Fig.\ref{fig:q1_q2_summary} support this view. Organization rankings vary considerably across domains and collaboration settings, and neither fixed nor learned organizations consistently dominate all conditions. The same communication graph can also produce different results when role assignment or input assignment changes. For example, under a fixed Tree graph, assigning a key role to a central node can be preferable in some settings, while assigning it to a bridge node becomes more effective in others. These results suggest that organization quality is not determined by the graph shape alone, but by how the communication graph, role assignment, and input assignment jointly match the collaboration required by the application.

Motivated by these findings, we present \textbf{OpenDGO}, a controlled benchmark and analysis framework for multi-agent organization. OpenDGO covers six application domains---Coding, Finance, Biomedical, Legal, Scientific Research, and Recommendation---and summarizes their collaboration requirements through three representative settings: \emph{Sequential Dependency}, \emph{Parallel Decomposition}, and \emph{Information Integration}. Under matched source tasks, agents, tools, and resource limits, OpenDGO compares six fixed organizations with recent learned organization methods. It further varies the communication graph, role assignment, input assignment, resource budget, and component failures separately, allowing the effect of each factor to be examined under the same evaluation setting. We also relate organization performance to graph properties and actual communication traces to explain why different organizations behave differently across applications. OpenDGO therefore aims to answer a concrete question: \emph{for a given application and collaboration requirement, which organization should be used, under what resource conditions, and why?}

\noindent\textbf{Our Contributions.}
\underline{\emph{(1) Comprehensive Benchmark.}}
We introduce OpenDGO, a controlled benchmark spanning six real-world domains, three collaboration settings, six fixed organizations, and recent learned organization methods, enabling organization evaluation across heterogeneous applications rather than within a single domain.
\underline{\emph{(2) Controlled Analysis.}}
We separately examine communication graph, role assignment, input assignment, resource budget, and failures, and connect these factors with graph properties and communication traces to explain their effects.
\underline{\emph{(3) Valuable Insights.}}
We show that no organization serves as a one-for-all solution: preferred organizations vary across applications, the same communication graph can behave differently under different role and input assignments, and practical choices involve clear quality--cost--robustness trade-offs.

\section{Preliminaries}
\label{sec:preliminaries}

\subsection{Notations and Organization Paradigms}
\label{sec:notations}

\textbf{Multi-Agent Tasks.}
Let $x$ denote a source problem from domain $d\in\mathcal{D}$, where $\mathcal{D}$ includes Coding, Finance, Biomedical, Legal, Scientific Research, and Recommendation. We denote its multi-agent form as $\mathcal{T}_{x,s}$, which preserves the original question $q$ and answer target while specifying the roles and inputs required for collaboration. Here, $s\in\mathcal{S}$ represents one of three collaboration settings:
\begin{equation}
    \mathcal{S}=\{\mathrm{Sequential},\mathrm{Parallel},\mathrm{Integration}\}.
\end{equation}
\emph{Sequential Dependency} requires agents to complete dependent steps in order; \emph{Parallel Decomposition} allows independent subtasks to be processed concurrently; and \emph{Information Integration} requires agents to combine and reconcile findings from different sources. These settings describe how agents need to collaborate without prescribing their organization.

\textbf{Multi-Agent Organizations.}
Given $\mathcal{T}_{x,s}$, an organization is defined as
\begin{equation}
    \mathcal{O}=(G,R,P,I,B),
\end{equation}
where $G=(V,E)$ is the communication graph, $R$ denotes agent roles, $P$ assigns roles to agents, $I$ assigns initial task inputs, and $B$ is the execution budget. Each vertex represents an agent, and edges specify permitted communication. At round $t$, agent $i$ produces
\begin{equation}
    h_i^{(t)}=\mathcal{M}\left(q,r_i,\iota_i,
    \{h_j^{(t-1)}:j\in\mathcal{N}_G(i)\};B\right),
    \label{eq:round_update}
\end{equation}
where $\mathcal{M}$ is the underlying LLM, $r_i$ and $\iota_i$ are the assigned role and task inputs, and $\mathcal{N}_G(i)$ denotes the communication neighbors. A designated agent combines the accessible outputs to produce the final answer.

\textbf{Organization Paradigms.}
We consider three paradigms. \emph{Single-Agent Reasoning} uses one agent with access to all task inputs. \emph{Fixed Organization} adopts a predefined graph, including Star, Chain, Tree, Ring, Mesh, and Community. \emph{Learned Organization} constructs task-dependent communication graphs, represented by G-Designer, ARG-Designer, and GoAgent~\cite{zhang2025gdesigner,li2026argdesigner,chen2026goagent}. These paradigms support comparisons under a unified evaluation setting.

\subsection{Related Work}
\label{sec:related_work}

Recent benchmarks have begun to evaluate how LLM agents collaborate rather than only whether they complete tasks. MultiAgentBench compares representative communication structures, including Star, Chain, and Tree~\cite{zhu2025multiagentbench}, while AgentsNet examines coordination across different network structures and agent scales~\cite{groetschla2025agentsnet}. More recent benchmarks investigate distributed coordination (SILO-BENCH), Web-agent collaboration (AgentWebBench), and collaboration under partial observability (MASBench)~\cite{zhang2026silobench,zhong2026agentwebbench,chu2026masbench}. However, these benchmarks mainly focus on specific environments or collaboration mechanisms, providing limited understanding of how organization preferences change across heterogeneous application domains and different role and input assignments. OpenDGO addresses this gap through controlled cross-domain comparisons to understand which organizations are effective under different application conditions and why.

\begin{figure*}[tp]
\centering
\makebox[\textwidth][r]{\makebox[0.66\textwidth][c]{%
  \includegraphics[width=0.40\textwidth,trim=88bp 91bp 74bp 0bp,clip]{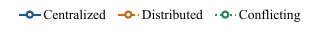}}}
\par\vspace{-27pt}
\begin{subfigure}[b]{0.34\textwidth}
  \centering
  \includegraphics[width=\linewidth]{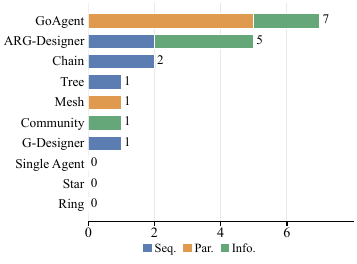}
  \caption{Organization wins.}
  \label{fig:q1_best_frequency}
\end{subfigure}
\hfill
\begin{subfigure}[b]{0.212\textwidth}
  \centering
  \includegraphics[width=\linewidth,trim=0bp 0bp 207.371bp 22bp,clip]{figures/q2_role_information_effects.pdf}
  \caption{Sequential tasks.}
  \label{fig:q2_sequential_allocation}
\end{subfigure}
\hfill
\begin{subfigure}[b]{0.212\textwidth}
  \centering
  \includegraphics[width=\linewidth,trim=115bp 0bp 103.371bp 22bp,clip]{figures/q2_role_information_effects.pdf}
  \caption{Parallel tasks.}
  \label{fig:q2_parallel_allocation}
\end{subfigure}
\hfill
\begin{subfigure}[b]{0.212\textwidth}
  \centering
  \includegraphics[width=\linewidth,trim=219bp 0bp 0bp 22bp,clip]{figures/q2_role_information_effects.pdf}
  \caption{Information Integration.}
  \label{fig:q2_information_allocation}
\end{subfigure}
\caption{\textbf{Organization preferences and role--input assignments.}
(a) Organization wins across 18 domain--task conditions.
(b--d) Effects of role and input assignments in a fixed Tree. Whiskers show intervals; C, B, and E denote Center, Bridge, and Edge.}
\label{fig:q1_q2_summary}
\end{figure*}

\section{Motivating Observations}
\label{sec:motivation}

Existing studies provide limited understanding of whether the same organization works across different application domains and how role and input assignments affect its performance. We examine these questions through two controlled analyses, as shown in Fig.~\ref{fig:q1_q2_summary}.

\textbf{Observation 1: Organization preference varies across applications.}
As shown in Fig.~\ref{fig:q1_best_frequency}, GoAgent and ARG-Designer win 12 of 18 domain--task conditions, while fixed organizations retain five wins. Among fixed organizations, Tree achieves the highest average quality on Sequential Dependency (62.6\%), whereas Community leads on Parallel Decomposition (67.0\%) and Information Integration (66.9\%). These results show that no organization consistently dominates across applications with different collaboration requirements.

\textbf{Observation 2: Communication graph alone is insufficient.}
We fix the six-agent Tree and vary role and input assignments. As shown in Fig.~\ref{fig:q2_sequential_allocation}--\ref{fig:q2_information_allocation}, Sequential tasks favor Center + Centralized (67.7\%), dropping to 60.2\% when the key role moves to Edge. Parallel Decomposition and Information Integration instead favor Bridge + Distributed (54.5\% and 58.7\%). Thus, organization effectiveness depends not only on the communication graph but also on how roles and inputs are assigned to agents.

\section{Organization Design and Analysis}
\label{sec:organization_analysis}

\begin{figure*}[t]
  \centering
  \includegraphics[width=\textwidth,pagebox=cropbox]{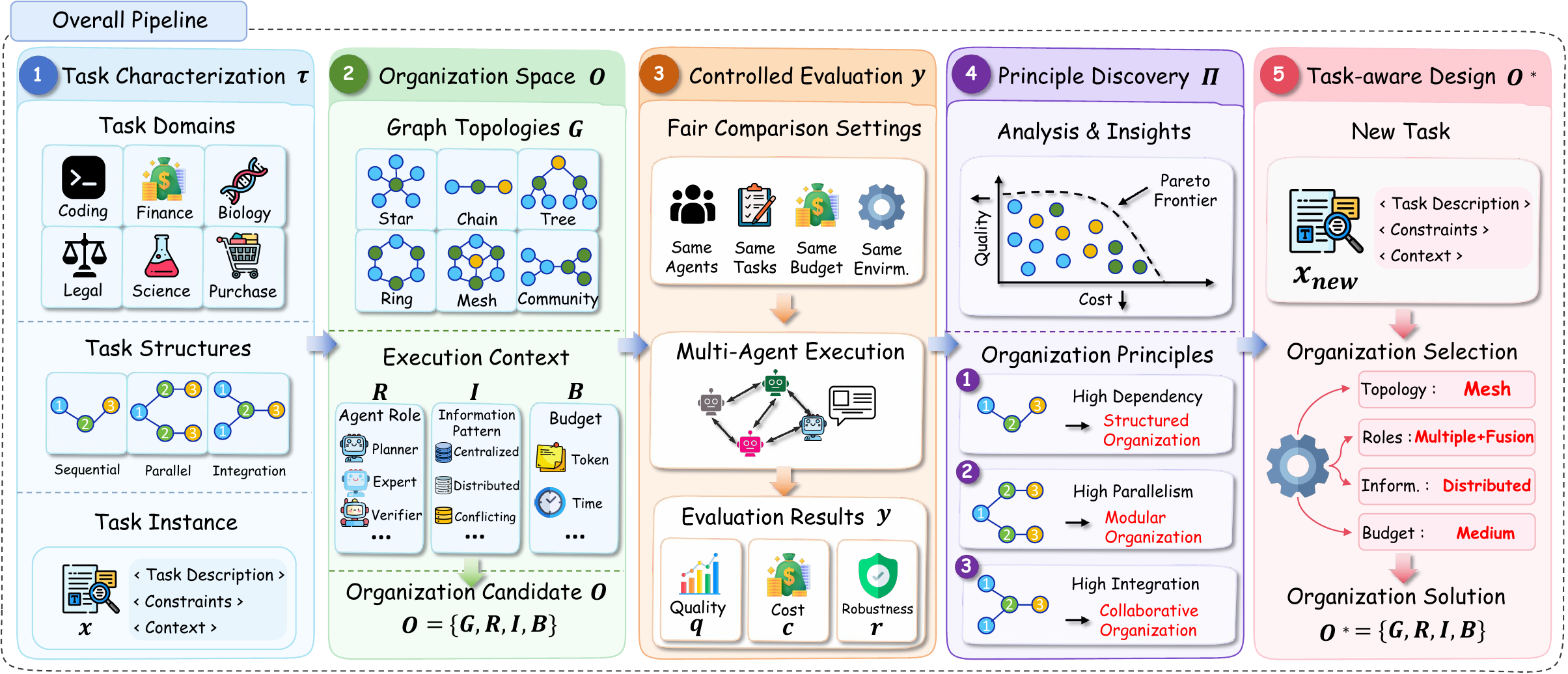}
  \caption{\textbf{Overview of the OpenDGO framework.} Task characterization,
  organization design, and controlled evaluation inform principle discovery
  and task-aware organization selection.}
  \Description{Five stages connect task characterization, the organization
  space, controlled evaluation, principle discovery, and task-aware design.
  The organization space contains graph topologies, agent roles, input
  assignments, and budgets. Evaluation compares task quality, cost, and
  robustness under matched conditions.}
  \label{fig:opendgo_framework}
\end{figure*}

OpenDGO aims to understand when a multi-agent organization is effective and why. We separate the collaboration required by a task from the organization used to support it, and relate controlled organization changes to communication behavior and task performance. Fig.~\ref{fig:opendgo_framework} summarizes the overall framework.

\subsection{Task Construction}
\label{sec:task_construction}

\textbf{Source Tasks.}
OpenDGO covers six domains: Coding, Finance, Biomedical, Legal, Scientific Research, and Recommendation. For each source instance, we preserve its original question, answer target, task inputs, and evaluator, while dividing the required work among six domain-specific roles with corresponding input assignments.

\textbf{Collaboration Structures.}
Each task is instantiated under three representative settings. \emph{Sequential Dependency} requires agents to complete dependent steps in order. \emph{Parallel Decomposition} distributes independent subtasks across agents before combining their outputs. \emph{Information Integration} requires agents to combine and reconcile information from different sources. These settings specify how agents collaborate without prescribing their communication graph.

\textbf{Task Dependencies.}
We represent the required collaboration as a task dependency graph $H_{x,s}$, whose vertices denote agent outputs and edges represent their dependencies. It specifies which outputs are needed by subsequent agents without determining how those agents communicate, allowing the same task requirements to be evaluated under different organizations.

\subsection{Controlled Organization Design}
\label{sec:organization_design}

\textbf{Organization Space.}
Following Sec.~\ref{sec:notations}, an organization $\mathcal{O}=(G,R,P,I,B)$ specifies the communication graph, agent roles, role assignment, input assignment, and execution budget. We consider six fixed organizations---Star, Chain, Tree, Ring, Mesh, and Community---together with three learned methods: G-Designer, ARG-Designer, and GoAgent. A Single-Agent setting with access to all task inputs serves as the reference.

\textbf{Controlled Interventions.}
OpenDGO examines organizational factors under matched source tasks, agent models, tools, and evaluation protocols. Varying $G$ compares communication graphs, while fixing $G$ and changing $P$ or $I$ isolates the effects of role and input assignments. We further examine graph properties through matched graph variants and evaluate different resource budgets and component failures. Other conditions are held fixed within each comparison.

\textbf{Execution Traces.}
Agents exchange messages only through permitted edges, and a designated agent produces the final answer from the inputs and messages available to it. Each execution records communication paths, intermediate outputs, message rounds, and token usage, allowing us to examine how different organizations operate.

\subsection{Graph-Based Collaboration Analysis}
\label{sec:graph_analysis}

To explain organization effectiveness, we examine three aspects of communication: information delivery, delivery resilience, and communication burden.

\textbf{Information Delivery.}
Let $\mathcal{U}^{*}$ denote the set of task inputs required for the final decision, and $A(u)$ the agents initially holding input $u$. The minimum number of communication rounds needed for all required inputs to reach the final decision agent is bounded below by
\begin{equation}
L_{\rm route}
=
\max_{u\in\mathcal{U}^{*}}
\min_{v\in A(u)}
\operatorname{dist}_{G}(v,v_{\rm syn}),
\label{eq:routing_bound}
\end{equation}
where $v_{\rm syn}$ denotes the final decision agent. This shows that delivery efficiency depends on both communication paths and input assignment.

\textbf{Delivery Resilience.}
Communication paths may become unavailable when agents or connections fail. For the agents $S^{*}$ that uniquely hold required inputs, we characterize their connection to the final decision agent by
\begin{equation}
\lambda^{*}
=
\min_{u\in S^{*}}
\lambda_G(u,v_{\rm syn}),
\label{eq:delivery_connectivity}
\end{equation}
where $\lambda_G$ denotes edge connectivity. More alternative paths can improve resilience against connection failures, although they cannot compensate for the failure of an agent holding a unique required input.

\textbf{Communication Burden.}
Additional connections may shorten communication paths but increase message processing. Assuming messages are sent to all permitted neighbors, the total message volume is
\begin{equation}
C_{\rm msg}
=
\sum_{t=0}^{T-1}
\sum_{i\in V}
\deg_G(i)\,
\ell(h_i^{(t)}),
\label{eq:message_exposure}
\end{equation}
where $\ell(h_i^{(t)})$ denotes the length of the message generated by agent $i$ at round $t$. Dense graphs may therefore reduce communication distance while increasing redundant messages and processing costs.

\textbf{Interaction Behavior.}
Graph properties describe possible communication paths, whereas execution traces reveal how agents actually communicate. We measure input coverage, propagation latency, duplicate-message rate, conflict resolution, cross-group communication, and final aggregation load to examine how organizational differences translate into task performance.

\subsection{Evaluation Protocol}
\label{sec:evaluation_protocol}

\textbf{Task Quality.}
Each domain retains its original evaluation metrics, including resolve rate, exact match or accuracy, claim-verification accuracy, and Hit@1 where appropriate.

\textbf{Coordination Behavior.}
Input coverage and propagation latency measure information delivery; duplicate-message rate and aggregation load capture communication overhead; conflict resolution and cross-group communication characterize collaboration between agents.

\textbf{Cost and Reliability.}
Token usage, message volume, and execution time measure organization cost. Controlled agent and communication-link failures evaluate the ability of different organizations to maintain performance under disruptions.

\begin{table*}[t]
\centering
\caption{\textsc{Organization benchmark (mean $\pm$ SD (\%)).}
\textbf{Bold} and \underline{underlining} mark the best and second-best result within each task-structure panel and domain.}
\label{tab:main_results}
\begingroup
\footnotesize
\setlength{\tabcolsep}{2.1pt}
\renewcommand{\arraystretch}{1.04}
\newcommand{\best}[2]{\textbf{#1}{\scriptsize$\pm$#2}}
\newcommand{\second}[2]{\underline{#1}{\scriptsize$\pm$#2}}
\newcommand{\qonebenchmarkrows}{%
\multicolumn{9}{c}{\cellcolor{opendgopanel}\bfseries (a) Sequential Dependency} \\
\cmidrule(lr){1-9}
\textbf{Single} & \cellcolor{opendgobaseline}Single Agent & 28.2{\scriptsize$\pm$1.9} & 52.6{\scriptsize$\pm$2.6} & 59.6{\scriptsize$\pm$2.5} & 65.2{\scriptsize$\pm$3.7} & 71.2{\scriptsize$\pm$1.1} & 37.2{\scriptsize$\pm$2.8} & 52.3{\scriptsize$\pm$1.4} \\
\cmidrule(lr){1-9}
\multirow[c]{6}{*}{\textbf{Fixed}} & \cellcolor{opendgofixed}Star & 24.8{\scriptsize$\pm$1.3} & 48.8{\scriptsize$\pm$3.1} & 55.8{\scriptsize$\pm$1.8} & 61.8{\scriptsize$\pm$3.0} & 67.6{\scriptsize$\pm$3.5} & 34.0{\scriptsize$\pm$2.8} & 48.8{\scriptsize$\pm$0.6} \\
& \cellcolor{opendgofixed}Chain & \best{37.6}{2.7} & \best{62.4}{5.8} & 67.6{\scriptsize$\pm$3.0} & 73.6{\scriptsize$\pm$3.2} & 78.4{\scriptsize$\pm$3.7} & 44.4{\scriptsize$\pm$4.9} & 60.7{\scriptsize$\pm$0.9} \\
& \cellcolor{opendgofixed}Tree & 34.2{\scriptsize$\pm$1.7} & 60.6{\scriptsize$\pm$3.0} & \best{71.2}{2.1} & 77.8{\scriptsize$\pm$2.6} & 81.2{\scriptsize$\pm$2.7} & 50.8{\scriptsize$\pm$3.8} & 62.6{\scriptsize$\pm$1.4} \\
& \cellcolor{opendgofixed}Ring & 29.4{\scriptsize$\pm$2.8} & 53.2{\scriptsize$\pm$4.5} & 61.0{\scriptsize$\pm$2.4} & 66.0{\scriptsize$\pm$3.1} & 71.8{\scriptsize$\pm$3.7} & 38.2{\scriptsize$\pm$4.3} & 53.3{\scriptsize$\pm$1.9} \\
& \cellcolor{opendgofixed}Mesh & 31.0{\scriptsize$\pm$2.5} & 56.8{\scriptsize$\pm$5.0} & 64.8{\scriptsize$\pm$2.9} & 70.2{\scriptsize$\pm$3.6} & 75.6{\scriptsize$\pm$4.1} & 42.6{\scriptsize$\pm$6.1} & 56.8{\scriptsize$\pm$1.8} \\
& \cellcolor{opendgofixed}Community & 35.8{\scriptsize$\pm$4.3} & 59.4{\scriptsize$\pm$4.8} & 69.4{\scriptsize$\pm$5.4} & 76.4{\scriptsize$\pm$4.1} & 79.8{\scriptsize$\pm$4.3} & 48.8{\scriptsize$\pm$5.1} & 61.6{\scriptsize$\pm$2.2} \\
\cmidrule(lr){1-9}
\multirow[c]{3}{*}{\textbf{Learned}} & \cellcolor{opendgolearned}G-Designer & 37.2{\scriptsize$\pm$4.2} & 62.0{\scriptsize$\pm$6.0} & \second{71.0}{2.8} & \second{78.2}{5.0} & \best{82.6}{6.9} & \second{51.2}{5.1} & \second{63.7}{2.5} \\
& \cellcolor{opendgolearned}ARG-Designer & \second{37.4}{5.1} & \second{62.2}{6.9} & 70.8{\scriptsize$\pm$3.5} & \best{79.0}{4.9} & \second{82.2}{6.1} & \best{52.0}{5.8} & \best{63.9}{1.8} \\
& \cellcolor{opendgolearned}GoAgent & 36.6{\scriptsize$\pm$3.7} & 61.4{\scriptsize$\pm$6.7} & 70.4{\scriptsize$\pm$5.5} & 77.6{\scriptsize$\pm$4.0} & 81.4{\scriptsize$\pm$4.1} & 50.4{\scriptsize$\pm$6.1} & 63.0{\scriptsize$\pm$1.6} \\
\midrule
\multicolumn{9}{c}{\cellcolor{opendgopanel}\bfseries (b) Parallel Decomposition} \\
\cmidrule(lr){1-9}
\textbf{Single} & \cellcolor{opendgobaseline}Single Agent & 27.4{\scriptsize$\pm$1.3} & 49.0{\scriptsize$\pm$2.6} & 55.2{\scriptsize$\pm$2.2} & 61.2{\scriptsize$\pm$2.8} & 69.8{\scriptsize$\pm$2.8} & 38.0{\scriptsize$\pm$4.6} & 50.1{\scriptsize$\pm$1.3} \\
\cmidrule(lr){1-9}
\multirow[c]{6}{*}{\textbf{Fixed}} & \cellcolor{opendgofixed}Star & 28.6{\scriptsize$\pm$3.7} & 50.4{\scriptsize$\pm$4.8} & 57.4{\scriptsize$\pm$2.2} & 64.0{\scriptsize$\pm$2.5} & 69.8{\scriptsize$\pm$2.5} & 41.6{\scriptsize$\pm$3.4} & 52.0{\scriptsize$\pm$1.5} \\
& \cellcolor{opendgofixed}Chain & 21.8{\scriptsize$\pm$3.6} & 43.0{\scriptsize$\pm$6.1} & 49.8{\scriptsize$\pm$2.1} & 55.8{\scriptsize$\pm$5.3} & 61.6{\scriptsize$\pm$4.1} & 29.0{\scriptsize$\pm$3.6} & 43.5{\scriptsize$\pm$1.1} \\
& \cellcolor{opendgofixed}Tree & 27.0{\scriptsize$\pm$2.8} & 49.2{\scriptsize$\pm$3.1} & 56.2{\scriptsize$\pm$3.1} & 62.2{\scriptsize$\pm$4.5} & 71.8{\scriptsize$\pm$2.8} & 36.6{\scriptsize$\pm$3.7} & 50.5{\scriptsize$\pm$1.3} \\
& \cellcolor{opendgofixed}Ring & 33.0{\scriptsize$\pm$3.5} & 57.0{\scriptsize$\pm$4.3} & 64.4{\scriptsize$\pm$3.3} & 68.8{\scriptsize$\pm$4.1} & 77.4{\scriptsize$\pm$4.8} & 46.2{\scriptsize$\pm$5.5} & 57.8{\scriptsize$\pm$1.0} \\
& \cellcolor{opendgofixed}Mesh & 41.6{\scriptsize$\pm$6.5} & 62.8{\scriptsize$\pm$5.3} & 71.8{\scriptsize$\pm$3.4} & 76.2{\scriptsize$\pm$4.3} & \best{84.0}{4.0} & 56.8{\scriptsize$\pm$5.7} & 65.5{\scriptsize$\pm$2.1} \\
& \cellcolor{opendgofixed}Community & 39.4{\scriptsize$\pm$3.3} & 64.6{\scriptsize$\pm$5.6} & 74.6{\scriptsize$\pm$3.7} & 79.4{\scriptsize$\pm$4.9} & 82.6{\scriptsize$\pm$4.8} & 61.6{\scriptsize$\pm$5.6} & 67.0{\scriptsize$\pm$1.4} \\
\cmidrule(lr){1-9}
\multirow[c]{3}{*}{\textbf{Learned}} & \cellcolor{opendgolearned}G-Designer & 42.2{\scriptsize$\pm$5.4} & 65.0{\scriptsize$\pm$6.8} & 75.2{\scriptsize$\pm$4.7} & 80.2{\scriptsize$\pm$5.3} & \second{83.8}{5.6} & 62.0{\scriptsize$\pm$6.6} & 68.1{\scriptsize$\pm$1.6} \\
& \cellcolor{opendgolearned}ARG-Designer & \second{42.8}{3.9} & \second{65.8}{5.1} & \second{76.0}{5.7} & \second{80.8}{4.6} & 83.6{\scriptsize$\pm$5.6} & \second{62.8}{5.0} & \second{68.6}{2.0} \\
& \cellcolor{opendgolearned}GoAgent & \best{43.6}{4.7} & \best{66.6}{6.3} & \best{76.6}{4.9} & \best{81.6}{6.2} & \second{83.8}{2.2} & \best{63.6}{5.8} & \best{69.3}{2.3} \\
\midrule
\multicolumn{9}{c}{\cellcolor{opendgopanel}\bfseries (c) Information Integration} \\
\cmidrule(lr){1-9}
\textbf{Single} & \cellcolor{opendgobaseline}Single Agent & 27.6{\scriptsize$\pm$2.8} & 49.8{\scriptsize$\pm$4.1} & 57.2{\scriptsize$\pm$2.0} & 62.2{\scriptsize$\pm$2.7} & 67.0{\scriptsize$\pm$2.9} & 35.2{\scriptsize$\pm$5.7} & 49.8{\scriptsize$\pm$0.6} \\
\cmidrule(lr){1-9}
\multirow[c]{6}{*}{\textbf{Fixed}} & \cellcolor{opendgofixed}Star & 25.2{\scriptsize$\pm$4.5} & 45.4{\scriptsize$\pm$4.4} & 52.4{\scriptsize$\pm$3.5} & 57.6{\scriptsize$\pm$4.0} & 61.2{\scriptsize$\pm$3.3} & 31.2{\scriptsize$\pm$5.7} & 45.5{\scriptsize$\pm$0.7} \\
& \cellcolor{opendgofixed}Chain & 23.2{\scriptsize$\pm$4.2} & 43.2{\scriptsize$\pm$5.4} & 50.2{\scriptsize$\pm$4.3} & 55.4{\scriptsize$\pm$3.0} & 59.2{\scriptsize$\pm$4.5} & 29.0{\scriptsize$\pm$5.3} & 43.4{\scriptsize$\pm$1.7} \\
& \cellcolor{opendgofixed}Tree & 30.4{\scriptsize$\pm$1.2} & 52.4{\scriptsize$\pm$4.9} & 60.4{\scriptsize$\pm$4.2} & 65.2{\scriptsize$\pm$1.8} & 69.6{\scriptsize$\pm$4.6} & 39.0{\scriptsize$\pm$6.2} & 52.8{\scriptsize$\pm$0.8} \\
& \cellcolor{opendgofixed}Ring & 33.2{\scriptsize$\pm$3.8} & 56.6{\scriptsize$\pm$5.7} & 65.2{\scriptsize$\pm$3.3} & 69.8{\scriptsize$\pm$4.6} & 74.8{\scriptsize$\pm$5.4} & 45.6{\scriptsize$\pm$4.7} & 57.5{\scriptsize$\pm$1.5} \\
& \cellcolor{opendgofixed}Mesh & 40.2{\scriptsize$\pm$4.2} & 63.4{\scriptsize$\pm$5.8} & 72.8{\scriptsize$\pm$4.8} & 77.8{\scriptsize$\pm$4.7} & 83.6{\scriptsize$\pm$3.8} & 55.8{\scriptsize$\pm$5.2} & 65.6{\scriptsize$\pm$1.6} \\
& \cellcolor{opendgofixed}Community & 41.0{\scriptsize$\pm$3.7} & 64.2{\scriptsize$\pm$5.4} & 74.2{\scriptsize$\pm$4.1} & 79.0{\scriptsize$\pm$4.3} & 83.0{\scriptsize$\pm$6.1} & \best{60.2}{7.1} & 66.9{\scriptsize$\pm$1.1} \\
\cmidrule(lr){1-9}
\multirow[c]{3}{*}{\textbf{Learned}} & \cellcolor{opendgolearned}G-Designer & 42.0{\scriptsize$\pm$5.5} & 65.4{\scriptsize$\pm$6.5} & 75.6{\scriptsize$\pm$5.1} & 80.6{\scriptsize$\pm$5.2} & 84.6{\scriptsize$\pm$5.9} & 59.8{\scriptsize$\pm$5.0} & \second{68.0}{2.4} \\
& \cellcolor{opendgolearned}ARG-Designer & \second{42.6}{5.5} & \best{66.0}{6.1} & \best{76.2}{6.1} & \best{81.2}{7.4} & \second{85.0}{5.3} & \second{60.0}{5.2} & \best{68.5}{2.8} \\
& \cellcolor{opendgolearned}GoAgent & \best{43.2}{6.0} & \second{65.6}{7.2} & \second{75.8}{5.8} & \second{80.8}{5.9} & \best{85.8}{6.2} & 59.6{\scriptsize$\pm$5.6} & \best{68.5}{0.8} \\
}

\begin{tabular*}{\textwidth}{@{\extracolsep{\fill}}llccccccc@{}}
\toprule
\multirow{2}{*}{\textbf{Category}} &
\multirow{2}{*}{\textbf{Method}} &
\multicolumn{7}{c}{\textbf{Task Quality (\%)}} \\
\cmidrule(lr){3-9}
& & \textbf{Code} & \textbf{Fin.} & \textbf{Bio.} & \textbf{Legal} &
\textbf{Sci.} & \textbf{Rec.} & \textbf{Avg.} \\
\midrule
\qonebenchmarkrows
\bottomrule
\end{tabular*}
\vspace{2pt}
\endgroup
\end{table*}

\section{Experiments and Analyses}
\label{sec:experiments}

We examine how organization choice depends on task demands, input placement,
and deployment constraints through four questions. \textbf{Q1:} How do
single-agent, fixed, and learned organizations compare across domains and task
structures? \textbf{Q2:} How do role placement and information allocation affect
a fixed topology? \textbf{Q3:} Which structural properties distinguish useful
organizations for different tasks? \textbf{Q4:} How does organization choice
change with token budgets, component failures, and unreliable inputs? These analyses progress from
comparative effectiveness to organizational factors, structural explanation,
and deployment trade-offs.

\subsection{Experimental Setup}
\label{sec:experimental_setup}

\textbf{Datasets and evaluation.}
We cover Coding with SWE-bench Verified~\citep{openai2024swebenchverified,jimenez2024swebench}, Finance with FinQA~\citep{chen2021finqa}, Biomedical with BioASQ Task B~\citep{tsatsaronis2015bioasq}, Legal with LegalBench~\citep{guha2023legalbench}, Scientific Research with SciFact~\citep{wadden2020scifact}, and Recommendation with Amazon Reviews 2023~\citep{hou2026amazon}. The domain-specific scores include resolve rate, exact match or accuracy, claim-verification accuracy, and Hit@1. Each domain covers Sequential Dependency, Parallel Decomposition, and Information Integration.

\textbf{Methods and comparison protocol.}
We compare Single Agent; six fixed organizations (Star, Chain, Tree, Ring, Mesh, and Community); and G-Designer~\citep{zhang2025gdesigner}, ARG-Designer~\citep{li2026argdesigner}, and GoAgent~\citep{chen2026goagent}. The single agent receives the complete set of task inputs available collectively to the six-agent systems. The target protocol pairs 500 tasks per domain--structure cell across methods, with the same model, tools, budget, stopping rule, scorer, and five seeds. Construction and execution details are in Appendix~\ref{app:methodology}.

\textbf{Supplementary analyses.}
We supplement Q1 with a backbone-transfer analysis and Q4 with theoretical cost accounting and an input perturbation stress test. Table~\ref{tab:backbone_organization_transfer} compares organization choices and transfer regret across backbones; Fig.~\ref{fig:q4_data_robustness} compares quality retention under four types of input perturbations. Table~\ref{tab:organization_complexity} in Appendix~\ref{app:organization_complexity} reports analytical costs rather than measured latency or memory.

\subsection{Organization Effectiveness (Q1)}
\label{sec:q1_benchmark}

To address \textbf{Q1}, we evaluate ten methods across six domains and three collaboration settings, and further examine whether organization preferences transfer across LLM backbones.

\begin{figure*}[tp]
  \centering
  \newsavebox{\opendgoPatternPanels}
  \newlength{\opendgoPatternBodyHeight}
  \sbox{\opendgoPatternPanels}{%
  \begin{minipage}[t]{0.66\textwidth}
  \vspace{0pt}
  \centering
  \begin{subfigure}[t]{0.37\linewidth}
    \centering
    \includegraphics[width=\linewidth]{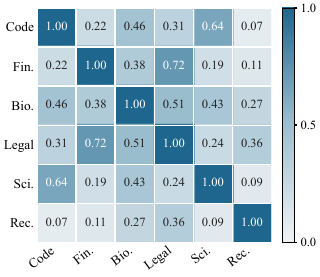}
    \caption{Domain-ranking similarity.}
    \label{fig:q1_domain_similarity}
  \end{subfigure}
  \hfill
  \begin{subfigure}[t]{0.61\linewidth}
    \centering
    \includegraphics[width=\linewidth]{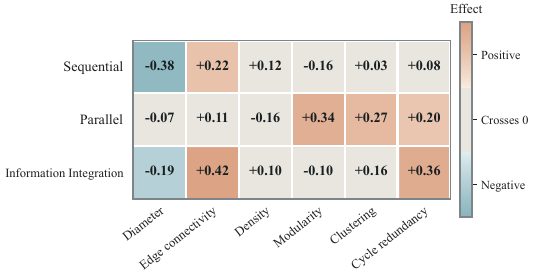}
    \caption{Graph-property effects.}
    \label{fig:q3_property_effects}
  \end{subfigure}
  \end{minipage}%
  }
  \setlength{\opendgoPatternBodyHeight}{%
    \dimexpr\ht\opendgoPatternPanels+\dp\opendgoPatternPanels\relax}
  \begin{minipage}[t]{0.66\textwidth}
  \vspace{0pt}
  \usebox{\opendgoPatternPanels}\par
  \captionsetup{justification=raggedright,singlelinecheck=false,position=bottom,skip=6pt}
  \caption{\textbf{Domain rankings and structural effects.} (a) Mean Kendall's $\tau_b$; (b) standardized partial effects on quality.}
  \label{fig:q1_q3_patterns}
  \end{minipage}
  \hfill
  \begin{minipage}[t]{0.32\textwidth}
    \vspace{0pt}
\begingroup
  \makeatletter
  \def\@captype{table}
  \makeatother
  \captionsetup{type=table,justification=raggedright,singlelinecheck=false,position=bottom,skip=6pt}
  \centering
  \begin{minipage}[t][\opendgoPatternBodyHeight][t]{\linewidth}
  \vspace{0pt}
  \fontsize{7}{8.4}\selectfont
  \setlength{\tabcolsep}{1pt}
  \makeatletter
  \dimen@=\dimexpr\opendgoPatternBodyHeight
    -2\heavyrulewidth-\lightrulewidth
    -2\aboverulesep-2\belowrulesep-\abovetopsep-\belowbottomsep\relax
  \divide\dimen@ by 4
  \@tempcnta=\dimen@
  \multiply\@tempcnta by 100
  \@tempcntb=\dimexpr\ht\strutbox+\dp\strutbox\relax
  \divide\@tempcnta by \@tempcntb
  \dimen@=\@tempcnta pt
  \divide\dimen@ by 100
  \edef\arraystretch{\strip@pt\dimen@}
  \makeatother
  \begin{tabular*}{\linewidth}{@{\extracolsep{\fill}}lcccc@{}}
    \toprule
    Backbone & \shortstack{Best\\Seq.} & \shortstack{Best\\Par.}
      & \shortstack{Best\\Info.} & \shortstack{Regret\\(pp) $\downarrow$} \\
    \midrule
   \shortstack[l]{Qwen3-8B\\-Instruct}
  & \shortstack{\textbf{Tree}\\(62.6)}
  & \shortstack{\textbf{Community}\\(67.0)}
  & \shortstack{\textbf{Community}\\(66.9)} & 0.00 \\

\shortstack[l]{Llama-3.1-8B\\-Instruct}
  & \shortstack{\textbf{Chain}\\(43.8)}
  & \shortstack{\textbf{Community}\\(48.6)}
  & \shortstack{\textbf{Mesh}\\(47.2)} & 0.43 \\

\shortstack[l]{Qwen2.5-1.5B\\-Instruct}
  & \shortstack{\textbf{Tree}\\(48.9)}
  & \shortstack{\textbf{Mesh}\\(54.1)}
  & \shortstack{\textbf{Community}\\(52.8)} & 0.20 \\
    \bottomrule
  \end{tabular*}
  \end{minipage}\par
  \caption{\textbf{Backbone choice and transfer regret.}}
  \label{tab:backbone_organization_transfer}
\endgroup

  \end{minipage}
\end{figure*}

\textbf{Overall Performance.}
Table~\ref{tab:main_results} shows that ARG-Designer leads on Sequential Dependency (63.9\%), GoAgent on Parallel Decomposition (69.3\%), and both on Information Integration (68.5\%), outperforming the best fixed organizations by 1.3, 2.3, and 1.6 percentage points, respectively. Among fixed organizations, Tree leads on Sequential tasks, while Community performs best on Parallel and Integration tasks. However, no organization dominates across domains: learned methods win 13 of 18 conditions (Fig.~\ref{fig:q1_best_frequency}), while domain-ranking similarities range from 0.07 to 0.72 (Fig.~\ref{fig:q1_domain_similarity}). These results highlight substantial variation in organization preferences across applications.

\textbf{Backbone Sensitivity and Transfer.}
We assess cross-backbone generalization using six fixed organizations under matched roles, inputs, tools, and execution policies. Let $\bar Q_{M,s}(o)$ denote the average quality of organization $o$ under backbone $M$ and setting $s$, and $o^*_{M,s}=\arg\max_o\bar Q_{M,s}(o)$. Transfer regret is defined as
\begin{equation}
\mathrm{Regret}(M_0\!\to\!M)
=\frac{1}{3}\sum_{s\in\mathcal S}
\bigl[\bar Q_{M,s}(o^*_{M,s})-\bar Q_{M,s}(o^*_{M_0,s})\bigr].
\label{eq:backbone_transfer_regret}
\end{equation}
Table~\ref{tab:backbone_organization_transfer} shows that preferred organizations vary across backbones, with transfer regrets of 0.43 and 0.20 percentage points for Llama and Qwen-2.5-32B, respectively. These modest losses suggest that organization preferences can change with model capabilities without substantial performance degradation. Details are provided in Appendix~\ref{app:backbone_transfer}.

\subsection{Role and Input Assignment (Q2)}
\label{sec:q2_factors}

To address \textbf{Q2}, we fix the six-agent Tree and vary expert placement (Center, Bridge, Edge) and input assignment (Centralized, Distributed, Conflicting), while keeping other settings unchanged.

\begin{figure*}[tp]
\centering
\includegraphics[width=\textwidth]{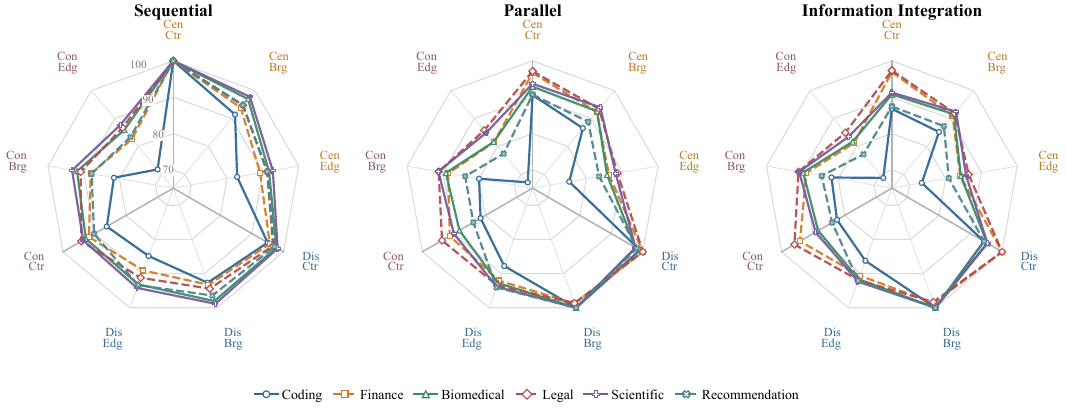}
\caption{\textbf{Domain-specific role--input configurations in a fixed Tree.} Each spoke combines input assignment (Cen: Centralized, Dis: Distributed, Con: Conflicting) and expert placement (Ctr: Center, Brg: Bridge, Edg: Edge).}
\label{fig:q2_configuration_map}
\end{figure*}

\textbf{Task-dependent configurations.}
Fig.~\ref{fig:q1_q2_summary}(b--d) shows that Sequential Dependency favors Center + Centralized (67.7\%), dropping to 60.2\% with Edge placement. Parallel Decomposition and Information Integration instead favor Bridge + Distributed (54.5\% and 58.7\%). These results indicate task-dependent preferences, although some intervals overlap.

\textbf{Domain variation.}
Fig.~\ref{fig:q2_configuration_map} compares nine configurations across domains. Center + Centralized leads on Sequential tasks, while Distributed inputs perform better on Parallel and Integration tasks. For the latter, Finance and Legal favor Center, whereas the other four domains favor Bridge. These findings highlight the joint importance of role and input assignments. Additional results and normalization details are provided in Appendices~\ref{app:q2_topologies} and~\ref{app:q2_domain_configurations}.

\subsection{Structure and Task Demands (Q3)}
\label{sec:q3_graph_properties}

To address \textbf{Q3}, we examine how graph properties relate to organization effectiveness. Fig.~\ref{fig:q1_q3_patterns} reports domain-ranking similarity and standardized structural associations across collaboration settings. Colored intervals indicate positive or negative associations, while gray intervals cross zero. Details are provided in Appendices~\ref{app:q3_details} and~\ref{app:q3_behavior}.

\begin{figure*}[tp]
  \centering
  \includegraphics[width=0.91\textwidth]{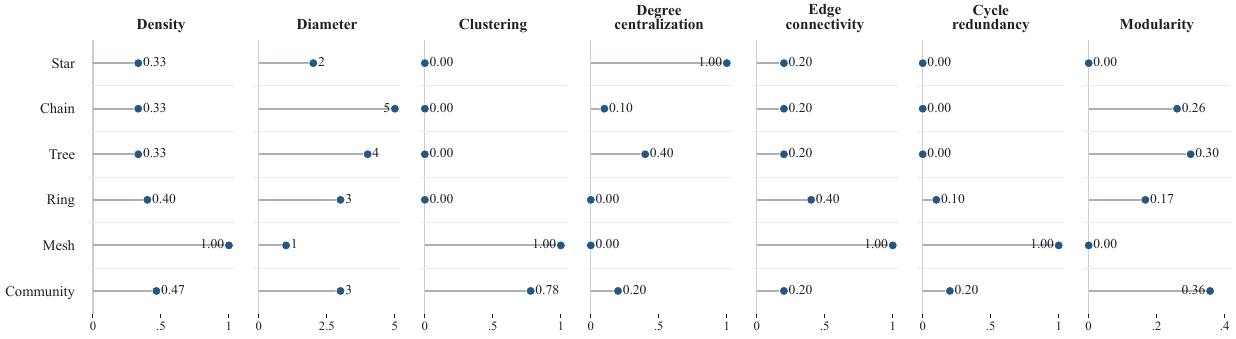}
  \caption{\textbf{Structural profiles of six organizations.} Markers show raw graph statistics on property-specific scales; larger values do not necessarily indicate better performance.}
  \label{fig:q3_property_atlas}
\end{figure*}

\textbf{Structural Differences.}
Fig.~\ref{fig:q3_property_atlas} shows that Star, Chain, and Tree share the same density (0.33) but differ in diameter (2, 5, and 4). Mesh minimizes communication distance, whereas Community exhibits stronger subgroup separation (modularity 0.36). Thus, graph density alone cannot explain organization effectiveness.

\textbf{Task-Specific Associations.}
As shown in Fig.~\ref{fig:q3_property_effects}, Sequential Dependency is associated with shorter diameter ($-0.38$) and higher connectivity ($+0.22$), favoring reliable information transfer. Parallel Decomposition shows positive associations with modularity ($+0.34$) and clustering ($+0.27$), supporting subgroup collaboration. Information Integration instead benefits from connectivity ($+0.42$) and cycle redundancy ($+0.36$), with a negative association for diameter ($-0.19$). These results reveal task-dependent structural preferences, although the associations do not establish causality.

\begin{figure*}[tp]
  \centering
  \includegraphics[width=0.94\textwidth]{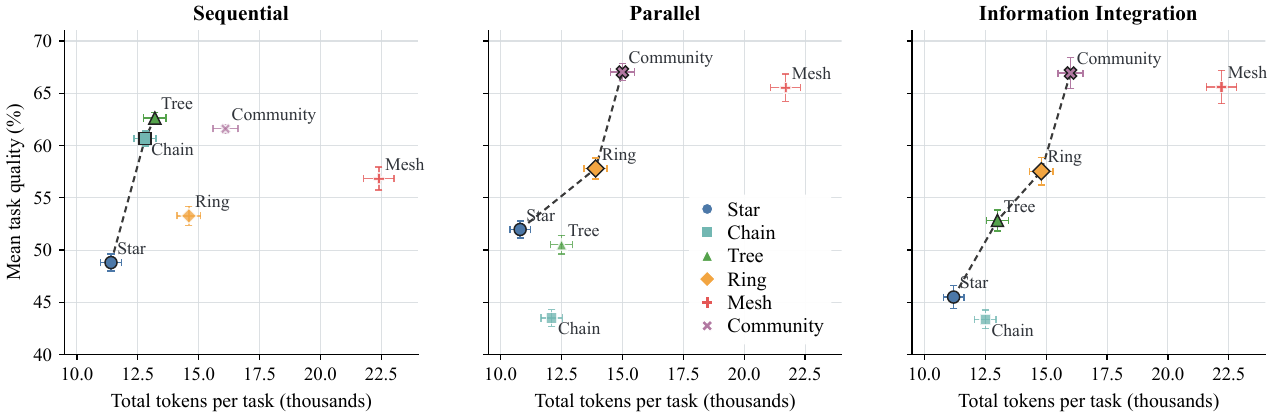}
  \caption{\textbf{Quality--token Pareto trade-offs.} Points show means, bars indicate one sample SD over five runs, and dashed lines connect non-dominated means.}
  \label{fig:q4_pareto}
\end{figure*}

\begin{figure*}[tp]
  \centering
  \includegraphics[width=0.94\textwidth]{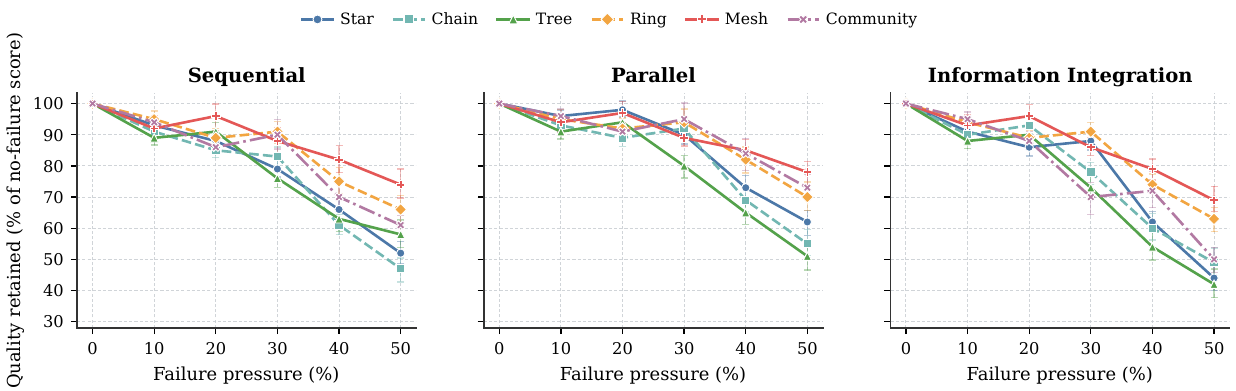}
  \caption{\textbf{Quality retained under component failures.} Results are normalized by no-failure quality; curves show means and bands indicate one sample SD.}
  \label{fig:q4_faults}
\end{figure*}

\begin{figure*}[tp]
  \centering
  \includegraphics[width=\textwidth]{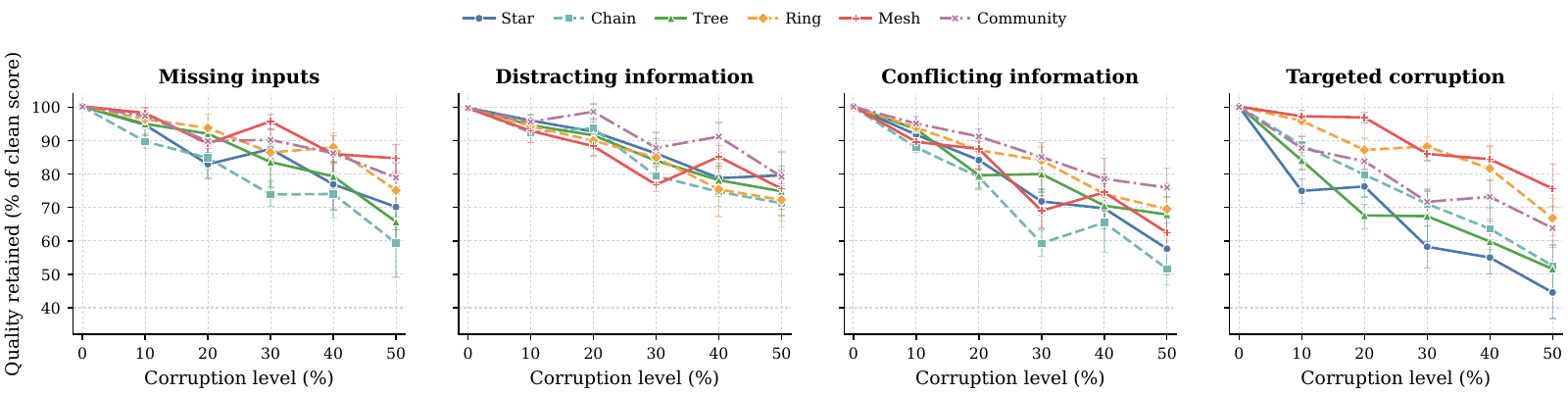}
  \caption{\textbf{Input robustness under Information Integration.} Four settings examine missing inputs, distracting information, conflicting information, and targeted input corruption. Curves show six-domain mean quality retention with 95\% bootstrap intervals.}
  \Description{Six fixed organizations under four input perturbations. Mesh performs better under missing inputs and targeted corruption, while Community is stronger against distracting and conflicting information.}
  \label{fig:q4_data_robustness}
\end{figure*}

\subsection{Quality, Cost, and Robustness (Q4)}
\label{sec:q4_tradeoffs}

To address \textbf{Q4}, we examine organization efficiency and robustness under different budgets, component failures, and input perturbations, with protocols detailed in Appendices~\ref{app:q4_details} and~\ref{app:q4_data_protocol}.

\textbf{Theoretical Costs.}
Table~\ref{tab:organization_complexity} in Appendix~\ref{app:organization_complexity} compares computation, communication, and memory costs. For $u$ agents, $v$ edges, $T$ rounds, average input length $b$, and message length $\ell$, the execution cost is approximated by
\begin{equation}
\OrgExec{u}{v}
=\Theta\!\left(uT\,\OrgCall{b+\frac{2v}{u}\ell}\right),
\label{eq:mas_execution}
\end{equation}
where $\OrgCall{s}$ denotes the cost of an LLM call with $s$ input tokens. Mesh has three times the edge-based message volume of Star, Chain, or Tree under matched rounds and message lengths, while learned organizations introduce additional graph-generation costs. Detailed complexity and memory analyses are provided in Appendix~\ref{app:organization_complexity}.

\textbf{Quality--Cost Trade-offs.}
Fig.~\ref{fig:q4_pareto} shows different Pareto-optimal organizations: Star--Chain--Tree for Sequential, Star--Ring--Community for Parallel, and Star--Tree--Ring--Community for Information Integration. Community achieves 67.0\% and 66.9\% on the latter two settings with 15--16k tokens, outperforming Mesh despite its higher cost of 21.7--22.4k tokens.

\textbf{Failure Robustness.}
Fig.~\ref{fig:q4_faults} shows that quality declines as agent or communication-link failures increase. At 50\% failure pressure, Mesh and Ring retain the highest relative quality, while Star, Chain, and Tree degrade more sharply. Thus, greater connectivity can improve robustness at the cost of additional communication.

\textbf{Input Robustness.}
We further test Information Integration under missing, distracting, conflicting, and targeted input perturbations, with corruption ratios $p\in\{0,0.1,0.2,0.3,0.4,0.5\}$. Quality retention is defined as
\begin{equation}
\mathrm{Ret}_{o,d,c}(p)
=100\,\frac{Q_{o,d,c}(p)}{Q_{o,d}(0)},
\label{eq:data_quality_retention}
\end{equation}
where $o$, $d$, and $c$ denote organization, domain, and perturbation type. As shown in Fig.~\ref{fig:q4_data_robustness}, Mesh retains 87.3\% under missing inputs and 81.0\% under targeted corruption at $p=0.5$, while Community leads against distracting and conflicting information (84.5\% and 75.8\%). These results suggest that communication redundancy improves tolerance to missing inputs but does not necessarily prevent the spread of incorrect information. Details are provided in Appendix~\ref{app:q4_data_protocol}.

\section{Conclusion and Future Directions}
\label{sec:conclusion}

In this work, we present \textbf{OpenDGO} to systematically study how LLM-based multi-agent systems should be organized. Across six real-world domains and three representative collaboration settings, we compare fixed and learned organizations and further examine role placement, information allocation, graph structure, resource budgets, and component failures. Our results show that organization effectiveness is strongly conditioned on the collaboration required by the task: no single topology consistently dominates, and the same graph can behave differently when roles or information are rearranged. Based on these findings, we highlight several directions for future multi-agent organization research.

\textbf{Task-Conditioned Organization.} Our results suggest that searching for a universally strong communication topology is unlikely to be sufficient. Sequential Dependency, Parallel Decomposition, and Information Integration tasks impose different information-flow requirements and therefore favor different structural properties. Future systems should infer collaboration demand from the task and adapt their organization accordingly, rather than applying the same topology to every instance.

\textbf{Organization Beyond Topology.} Communication structure alone does not determine team effectiveness. Role placement and information allocation can substantially change the behavior of the same graph, indicating that topology, capability placement, and information distribution should be considered jointly. Future organization designers should therefore move beyond optimizing only $G$ toward jointly reasoning over $(G,P,I)$ and, when necessary, resource constraints $B$.

\textbf{Behavior-Grounded Organization Design.} Static graph statistics provide only an incomplete explanation of multi-agent collaboration. Our analysis shows that structural properties matter through their effects on information delivery, propagation delay, redundancy, and information integration during execution. Future methods could directly use message-level traces and information-flow signals to design organizations, so that communication structures are optimized for the collaboration they actually induce.

\textbf{Resource-Aware Adaptive Organization.} Higher connectivity can improve information access and failure tolerance, but it also increases communication and synthesis costs. The preferred organization may therefore change with token budgets, latency requirements, or component availability. A promising direction is dynamic organization, where communication remains sparse when local reasoning is sufficient, expands when distributed information must be integrated, and reconfigures when agents or communication links become unavailable.

\clearpage
\makeatletter
\global\@colroom=\textheight
\global\vsize=\textheight
\makeatother

\bibliographystyle{ACM-Reference-Format}
\bibliography{references}

\newpage
\appendix
\section{Methodology in Detail}
\label{app:methodology}

\subsection{Benchmark Unit and Estimand}
\label{app:benchmark_unit}

For organization $a$, let $Y(T_{x,s},a,M,z)\in[0,1]$ be the domain scorer's
output for source task $x$, structural view $s$, shared model $M$, and seed
$z$. The paired contrast against organization $b$ is
\begin{equation}
\widehat\Delta_{d,s}(a,b)=\frac{1}{|\mathcal X_{d,s}|K}
\sum_{x\in\mathcal X_{d,s}}\sum_{z=1}^{K}
\bigl[Y(T_{x,s},a,M,z)-Y(T_{x,s},b,M,z)\bigr].
\label{eq:paired_effect}
\end{equation}
Pairs share the task, seed, combined inputs, model, tools, scorer, and nominal
limits. Source-ID cluster resampling preserves all paired configurations and
structural views. Comparisons of native frameworks additionally vary their
execution policies; these are system comparisons rather than topology-only
contrasts. A framework enters the controlled graph comparison only if its
role set, input boundary, scheduling, and synthesis policy can be matched.

\subsection{Task Construction and Structural Matching}
\label{app:task_construction}

Each package records the dataset version, split, instance and source IDs,
selection rule, and a hash of frozen model-visible input. Gold answers,
hidden tests, and support labels remain evaluator-only. Where redistribution
is restricted, IDs, conversion rules, and hashes provide the reconstruction
record. Selection and conversion precede held-out scoring.

Sequential packages specify ordered intermediate products; Parallel packages
specify independently analyzable views and their integration; Information Integration
packages partition complementary units and, where applicable, verified
conflicting or distracting units. The imposed dependency graph describes the
conversion, not the source dataset's original purpose. A conflicting view
retains a valid answer path. Packages are screened for an automatic target and
sufficient information in the union of their views. Difficulty stratum, input
length, input-unit count, intermediate decisions, and source category are
recorded. Cross-structure comparisons use matched sources where possible;
different source compositions are recorded otherwise.

\begin{table*}[t]
\centering
\caption{Domain-specific realizations of the three task structures. Each row
starts from a source problem with a fixed answer target.}
\label{tab:structure_instantiations}
\begingroup\small\setlength{\tabcolsep}{3pt}\renewcommand{\arraystretch}{1.12}
\begin{tabular}{>{\raggedright\arraybackslash}p{0.15\textwidth}>{\raggedright\arraybackslash}p{0.24\textwidth}>{\raggedright\arraybackslash}p{0.24\textwidth}>{\raggedright\arraybackslash}p{0.31\textwidth}}
\toprule
Domain & Sequential Dependency & Parallel Decomposition & Information Integration \\
\midrule
Coding & Locate, patch, test & Inspect files or failures separately & Reconcile issue, code, and test results \\
Finance & Extract figures, compute, verify & Inspect table and narrative separately & Combine complementary numerical facts \\
Biomedical & Retrieve, extract, verify & Review distinct articles & Resolve complementary or conflicting snippets \\
Legal & Identify issue, apply rule & Review facts, rules, exceptions & Reconcile rule, facts, and counterarguments \\
Scientific & Retrieve, classify, check rationale & Review candidate abstracts & Weigh support against refutation \\
Rec. & Profile, filter, rank & Score candidates by distinct signals & Combine history, reviews, and metadata \\
\bottomrule
\end{tabular}
\endgroup
\end{table*}

Role responsibilities are domain-specific: Coding separates issue analysis,
navigation, implementation, testing, regression review, and integration;
Finance separates question, table, and text analysis, calculation,
verification, and integration; Biomedical separates question typing,
retrieval, extraction, reasoning, fact checking, and integration. Legal uses
issue, fact, and rule analysis, exception review, counterargument review, and
integration; Scientific Research uses claim analysis, retrieval, support and
refutation analysis, rationale checking, and integration; Recommendation uses
profiling, review and metadata analysis, scoring, constraint review, and
integration. Private views determine access independently of role titles.

\subsection{Six-Agent Communication Graphs}
\label{app:fixed_graphs}

All fixed organizations use $V=\{0,1,2,3,4,5\}$. An undirected edge permits
messages in both directions. Table~\ref{tab:graph_edges} fixes the edge sets;
graph labels do not imply different prompts or models.

\begin{table}[t]
\centering
\caption{Exact fixed six-vertex graphs. Edge $ij$ joins vertices $i$ and $j$.}
\label{tab:graph_edges}
\small
\begin{tabular}{lll}
\toprule
Graph & Edges & $|E|$ \\
\midrule
Star & $01,02,03,04,05$ & 5 \\
Chain & $01,12,23,34,45$ & 5 \\
Tree & $01,02,13,14,25$ & 5 \\
Ring & $01,12,23,34,45,50$ & 6 \\
Mesh & all $\binom{6}{2}$ pairs & 15 \\
Community & $01,02,12,34,35,45,23$ & 7 \\
\bottomrule
\end{tabular}
\end{table}

The Community graph has two triangles joined by bridge $23$. Synthesizer
location and scheduling are held fixed within paired comparisons and logged
with the graph ID. Learned or native systems record their realized routes and
policy differences rather than inheriting the fixed-graph assumptions.

\subsection{Execution and Information Boundaries}
\label{app:execution}

Equation~\ref{eq:round_update} applies after initialization
$h_i^{(0)}=F_M(q,r_i,e_i,\varnothing;B)$. Messages expose only neighbor outputs;
there is no unrestricted transcript at synthesis. Logs identify sender,
recipient, round, delivered content or its hash, and token usage. Tool events
and shared state are recorded separately. Coding uses a frozen repository
snapshot and submits a patch to the official evaluator; any shared workspace
access is outside the pure message-routing bound.

Final answers use a frozen schema and evaluator. Token accounting sums actual
model inputs and outputs, including tool responses and final synthesis;
received-message tokens already counted as inputs are not added again.
Routed-token exposure in Eq.~\ref{eq:message_exposure} is a separate diagnostic.
Invalid attempted outputs remain in the scoring denominator under the domain
scorer. Prompt, parser, truncation, and stopping choices are fixed before
held-out comparisons.

\subsection{Controlled Intervention Details}
\label{app:interventions}

For Q2, the Tree places the designated expert at Center (vertex 0), Bridge
(1 or 2), or Edge (3, 4, or 5), with a compensating role permutation.
The selected vertex and synthesizer are recorded because members of a named
class can have different distances. Centralized allocation bundles decisive
units at a designated holder; Distributed allocation separates complementary
units; Conflicting allocation separates gold-supporting and verified
contradictory or distracting units. Inventory and target remain fixed, with
unit source, destination, and truth status recorded.

For mean score $\bar Y_{d,s}(p,i)$, a position contrast holds allocation fixed;
an allocation contrast holds position fixed. Their interaction is
\begin{equation}
\begin{aligned}
\Gamma_{d,s}={}&[\bar Y_{d,s}(C,Cen)-\bar Y_{d,s}(E,Cen)]\\
&-[\bar Y_{d,s}(C,Dis)-\bar Y_{d,s}(E,Dis)],
\end{aligned}
\label{eq:placement_information_interaction}
\end{equation}
where $C/E$ mean Center/Edge and $Cen/Dis$ mean Centralized/Distributed.
Non-Tree position classes and symmetry controls are specified in
Appendix~\ref{app:q2_topologies}. Q3 graph models and trace definitions appear
in Appendices~\ref{app:q3_details} and~\ref{app:q3_behavior}; Q4 caps and fault
sampling appear in Appendix~\ref{app:q4_details}. Each fault record identifies
the disabled component and its relationship to required sources and routes.

\subsection{Derivations and Scope of the Graph Perspectives}
\label{app:graph_lens_derivations}

\textbf{Routing bound.} Consider a support unit absent from the public
question, shared context, and independently available tools. At step zero,
its source-dependent information is available only to $A(e)$. Inductively,
after $t$ exchanges it can influence only vertices within distance $t$ of
$A(e)$, since each update reads neighbors' preceding states. Its provenance
cannot reach $v_{\rm syn}$ before $\min_{u\in A(e)}\operatorname{dist}_G(u,v_{\rm syn})$.
Taking the maximum across designated units gives Eq.~\ref{eq:routing_bound};
an unreachable unit has infinite distance. With replicated inputs the
minimum is over all holders. The argument gives an earliest possible arrival,
not a guarantee of faithful forwarding, conflict resolution, or correct
inference. A final synthesis that reads neighbors counts as an additional
exchange step when reporting this bound.

\textbf{Critical-cut condition.} For surviving source and synthesis vertices,
a cut of size $\lambda_G(u,v_{\rm syn})$ can disconnect their information path.
Fewer deletions cannot disconnect it. Taking the minimum across uniquely held
required sources proves the stated condition $k<\lambda^*$. Menger's theorem
provides the equivalent edge-disjoint-path characterization
\citep{diestel2025graph}. This protects connectivity under edge deletions;
internal-vertex failures require internally vertex-disjoint paths and endpoint
failures require input or decision redundancy. For example, Community has
cycles within both subgroups but a single bridge between them. Information that
must cross that bridge is unprotected against its deletion regardless of
those local cycles.

\textbf{Communication exposure.} At each exchange, vertex $i$ sends its
$\ell(h_i^{(t)})$ tokens to $\deg_G(i)$ recipients. Summing gives
Eq.~\ref{eq:message_exposure}, with $2T|E|$ directed deliveries if every
permitted message is sent. Dropped or truncated messages reduce exposure;
asynchronous or selective routing requires summing realized deliveries
instead. Equal output lengths would make exposure proportional to edge
count, but output lengths and reasoning quality can change with organization.
Consequently, density supplies a communication opportunity measure rather
than a sufficient predictor of cost or performance.

\section{Additional Experimental Details}
\label{app:experimental_details}

\subsection{Organization Generality and Domain Sensitivity}
\label{app:q1_generality}

The Q1 diagnostics are displayed in the main paper:
Fig.~\ref{fig:q1_best_frequency} complements the domain--structure results in
Table~\ref{tab:main_results}. GoAgent wins seven conditions, ARG-Designer five,
and fixed graphs five; G-Designer accounts for the remaining win.
Fig.~\ref{fig:q1_domain_similarity} orders the six fixed organizations by
domain and task/seed and averages Kendall's $\tau_b$ coefficients across the
15 domain pairs.
Finance--Legal has the highest displayed similarity ($0.72$), whereas
Coding--Recommendation ($0.07$), Scientific--Recommendation ($0.09$), and
Finance--Recommendation ($0.11$) have the lowest. These summaries do not
substitute for an evaluation of transfer to unseen tasks.

\subsection{Backbone-Transfer Protocol}
\label{app:backbone_transfer}

The comparison in Table~\ref{tab:backbone_organization_transfer}
changes only the execution backbone: task IDs, input inventory, six-node
graph, role assignment, information allocation, synthesizer, tools,
communication rounds, and scoring remain fixed. Shared word or character
limits for messages and final answers avoid equating different tokenizers;
actual token use is recorded separately. Exact checkpoint revisions,
decoding and thinking modes, and failure/retry policies must be logged.

The best fixed organization is selected separately for each task structure
using the unweighted six-domain average. Single Agent and learned designers
are excluded from that argmax. Source-backbone choices should be frozen on
a development set before target-backbone testing. Equation~\ref{eq:backbone_transfer_regret}
then compares the target model's own best graph with the transferred graph
on the same held-out tasks. Target best-graph selection should use a
separate selection split where possible; otherwise its test-set argmax is
an optimistic oracle reference and must be reported as such. Raw quality,
per-structure regret, and paired intervals are needed alongside the mean.
Source-ID cluster bootstrap retains all backbones, graphs, and seeds of a
source together. Close mean scores without a paired interval do not justify
a claim of a reliably different winner.

For the Llama row, the per-structure losses are
$0.4$, $0.0$, and $0.9$ pp, yielding $(0.4+0+0.9)/3=0.43$ pp. The
Qwen-2.5-32B row has losses $0.0$, $0.6$, and $0.0$ pp, yielding
$0.20$ pp. The three task structures receive equal weight in the mean.

\subsection{Normalization of the Tree Configuration Profiles}
\label{app:q2_domain_configurations}

To compare configuration sensitivity across domains with different score
levels, Fig.~\ref{fig:q2_configuration_map} normalizes each domain--task
profile separately. For configuration $c$ among the nine role--information
combinations, the displayed value is
$100\,\bar Y_{d,s}(c)/\max_{c'}\bar Y_{d,s}(c')$, where $\bar Y$ is the
five-seed mean. The highest value in each profile is therefore 100, and all
within-profile orderings are preserved. The radial scale starts at 65.
Curve area and values across domains are not measures of absolute task
quality. These profiles display means without uncertainty intervals and do
not establish significant differences between nearby configurations.

\subsection{Role--Information Interventions Beyond Tree}
\label{app:q2_topologies}

The Q2 intervention is repeated on the five remaining fixed topologies with
the task set, seeds, roles, input inventory, prompts, and budget held fixed.
Positions are defined by structural equivalence: Star compares Hub and Leaf;
Chain compares Center, Relay, and End; Community compares Core, Bridge, and
Peripheral. All vertices of Ring and Mesh are automorphism-equivalent, so
their displayed rotations or node sets are repeated placement controls rather
than distinct centrality classes. Information remains Centralized,
Distributed, or Conflicting under the same construction used for Tree.

\begin{figure*}[tp]
\centering
\begin{subfigure}[t]{0.98\textwidth}
  \centering
  \includegraphics[width=\linewidth]{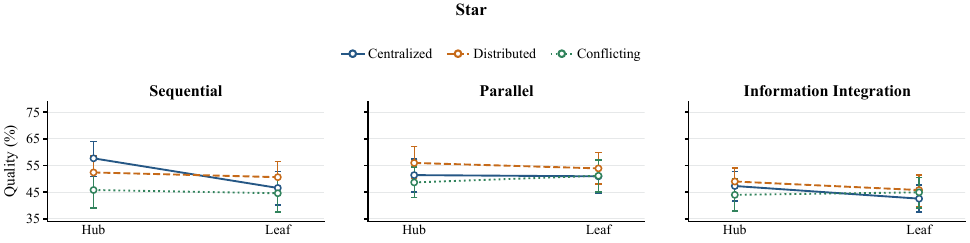}
  \caption{Star: hub versus leaf placement.}
\end{subfigure}

\vspace{5pt}
\begin{subfigure}[t]{0.98\textwidth}
  \centering
  \includegraphics[width=\linewidth]{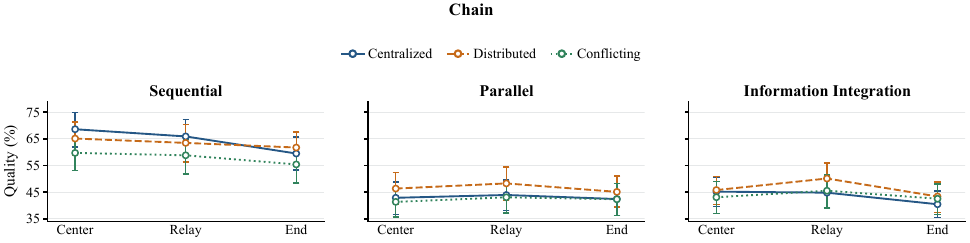}
  \caption{Chain: center, relay, and end placement.}
\end{subfigure}
\caption{\textbf{Q2 extension to sparse non-Tree topologies.} Curves show the
three information allocations; whiskers show intervals.}
\label{fig:q2_appendix_sparse}
\end{figure*}

\begin{figure*}[tp]
\centering
\begin{subfigure}[t]{0.98\textwidth}
  \centering
  \includegraphics[width=\linewidth]{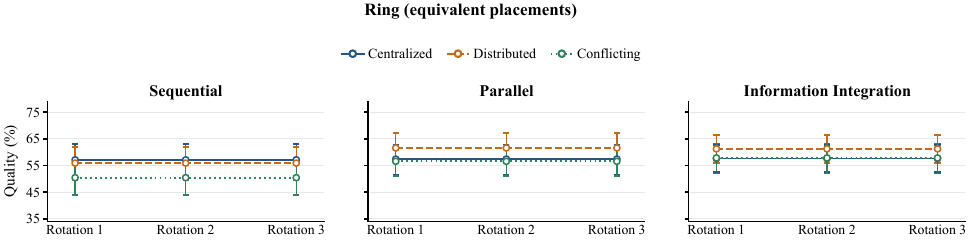}
  \caption{Ring: the three rotations are structurally equivalent controls.}
\end{subfigure}

\begin{subfigure}[t]{0.98\textwidth}
  \centering
  \includegraphics[width=\linewidth]{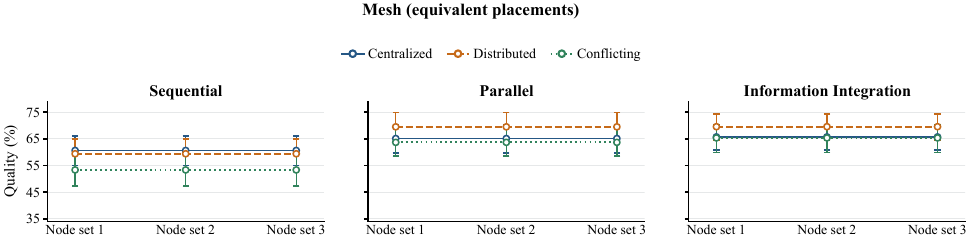}
  \caption{Mesh: the three node sets are structurally equivalent controls.}
\end{subfigure}

\begin{subfigure}[t]{0.98\textwidth}
  \centering
  \includegraphics[width=\linewidth]{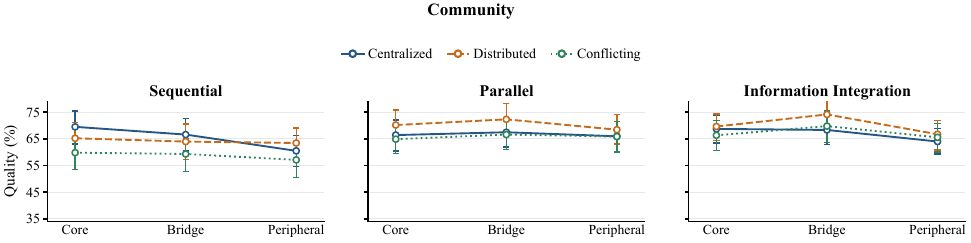}
  \caption{Community: core, bridge, and peripheral placement.}
\end{subfigure}
\caption{\textbf{Q2 extension to cyclic, dense, and modular topologies.}
Curve and interval semantics match Fig.~\ref{fig:q2_appendix_sparse}}
\label{fig:q2_appendix_dense}
\end{figure*}

\subsection{Controlled Graph-property Analysis}
\label{app:q3_details}

For each six-node organization, density measures communication volume;
diameter measures worst-case propagation distance; mean clustering measures
local cohesion; Freeman degree centralization measures hub dependence;
normalized edge connectivity measures disconnection resistance; normalized
independent-cycle redundancy measures alternative routing; and greedy
modularity measures subgroup separation.

Graph statistics cannot generally be varied independently. We therefore build
matched connected six-node variants that alter one focal property while
minimizing changes in the others. The same six focal properties---diameter,
edge connectivity, density, modularity, clustering, and cycle redundancy---are
to be evaluated under every task structure. Within each task, a separate
standardized partial-effect model will be fitted for each property, adjusting for
residual changes in other graph properties and domain fixed effects while
agents, roles, input allocation, inputs, and budget remain fixed.

For a connected six-node graph with $m$ edges, density is $m/15$ and normalized
cycle redundancy is $(m-5)/10$, so redundancy equals
$1.5\,\mathrm{density}-0.5$. Separate focal-property models must omit this
collinear counterpart from their controls; the two coefficients cannot be
interpreted as independently identified effects of adding edges. Constant or
singular graph controls must likewise be excluded. The message-trace measures
below provide diagnostics for the proposed mechanisms.

\subsection{Interaction-Behavior Measures}
\label{app:q3_behavior}

To organize the explanation of quality differences, the analysis follows
\begin{equation}
\phi(G)\ \longrightarrow\ \mathbf m(T_{x,s},\mathcal O)
\ \longrightarrow\ Y(T_{x,s},\mathcal O,M,z),
\label{eq:mechanism_chain}
\end{equation}
where $\phi(G)$ contains graph properties, $\mathbf m$ contains observed
interaction measures, and $Y$ is the domain score. This is an analytical
sequence linking structure, behavior, and outcome, not an identified causal
chain. Analyses within domains retain task structure; pooled analyses include
domain differences and assess domain interactions where identifiable.

Graph statistics describe communication opportunities, whereas message traces
show which opportunities are used. We therefore compute the six mediators in
Table~\ref{tab:interaction_behavior} from delivered messages, input-unit IDs,
and final-synthesis events. All text normalization, similarity thresholds, and
denominators are fixed before comparing organizations. Runs with no applicable
conflict or community partition are marked not applicable for the relevant
metric rather than assigned a favorable zero.

\begin{table*}[tp]
\centering
\caption{Q3 interaction measures. $\mathcal E^*$: designated supporting
information; $R$: round cap.}
\label{tab:interaction_behavior}
\small
\setlength{\tabcolsep}{4pt}
\begin{tabular}{>{\raggedright\arraybackslash}p{0.19\textwidth}>{\raggedright\arraybackslash}p{0.51\textwidth}>{\raggedright\arraybackslash}p{0.22\textwidth}}
\toprule
Measure & Operational definition & Interpretation \\
\midrule
Information coverage & $|\mathcal E^*\cap\mathcal E_{\rm syn}|/|\mathcal E^*|$ at final synthesis & Required information that reaches the decision point \\
Propagation latency & First round $t$ when all designated support units reach the synthesizer; incomplete cases are censored at $R+1$ & Delay introduced by routing paths \\
Duplicate-message rate & Fraction of delivered messages with token-set Jaccard similarity at least $0.90$ to an earlier message in the run & Repetition and communication waste \\
Contradiction resolution & Source-grounded resolved conflicts divided by injected, answer-relevant conflicts & Ability to reconcile conflicting information \\
Cross-group flow & Information-bearing messages crossing the fixed community cut divided by all information-bearing messages & Exchange between specialized groups \\
Synthesizer load & Input and received-message tokens processed in the final synthesis event & Aggregation burden at the decision point \\
\bottomrule
\end{tabular}
\end{table*}

For each task structure, we first regress a standardized behavior measure on a
standardized graph property with domain fixed effects and the matched-variant
controls used in Appendix~\ref{app:q3_details}. We then add that behavior
measure to the corresponding quality model. We report the property--behavior
coefficient, the behavior--quality coefficient conditional on the property,
and the change in the property coefficient. Bootstrap resampling keeps every
task's organizations and seeds together. This analysis diagnoses a proposed
mechanism; correlated graph properties and trace measures preclude a claim of
fully identified causal mediation.

\subsection{Budget and Fault Protocols}
\label{app:q4_details}

\textbf{Token accounting.}
We count agent inputs, outputs, inter-agent messages, and final aggregation. An
organization is Pareto-efficient if no alternative has no greater mean token
cost and no lower mean quality, with at least one strict improvement.
Common per-task caps range from 6k to 24k tokens
and reserve the same final-answer allowance. When collaboration budget is
exhausted, later messages are suppressed and the synthesizer uses the state
already received. Caps are enforced at message boundaries, so crossing a
threshold may enable a complete additional round and need not yield a smooth
curve.

\textbf{Fault injection.}
At pressure $p\in\{0,10,20,30,40,50\}\%$, each trial selects agent or active
message-channel failure with equal probability and disables proportion $p$
before execution. Failed agents emit no output and failed channels drop messages
without retry. Locations vary by seed, so damage depends on whether an
integrator, articulation point, or redundant component is sampled. Quality is
normalized by each topology's no-failure score to separate resilience from
baseline accuracy.

\subsection{Organization-Cost Definitions and Assumptions}
\label{app:organization_complexity}

\begin{table*}[t]
  \centering
  \caption{\textbf{Theoretical costs of representative organizations.}
  Per-task analysis assumes undirected routing and retained interaction
  traces. $\mathcal X$ and $\mathcal S$ summarize execution and trace state;
  $\mathcal F_i$ counts offline designer fitting only. Shared backbone
  parameters, KV cache, backbone training, and supervision rollouts are excluded.}
  \label{tab:organization_complexity}
  \begingroup
  \footnotesize
  \setlength{\tabcolsep}{3pt}
  \renewcommand{\arraystretch}{1.15}
  \resizebox{\textwidth}{!}{%
  \begin{tabular}{@{}lccccc@{}}
    \toprule
    Organization & Active edges & \shortstack{Offline\\designer fitting}
      & Online cost per task & Routed message tokens & Runtime state \\
    \midrule
    Single Agent & $0$ & -- & $\Theta(\OrgCall{b_{\rm all}})$
      & $0$ & $\Theta(b_{\rm all})$ \\
    \midrule
    Star & $n-1$ & -- & $\OrgExec{n}{n-1}$
      & $\Theta(T(n-1)\ell)$ & $\OrgState{n}{n-1}$ \\
    Chain & $n-1$ & -- & $\OrgExec{n}{n-1}$
      & $\Theta(T(n-1)\ell)$ & $\OrgState{n}{n-1}$ \\
    Tree & $n-1$ & -- & $\OrgExec{n}{n-1}$
      & $\Theta(T(n-1)\ell)$ & $\OrgState{n}{n-1}$ \\
    Ring & $n$ & -- & $\OrgExec{n}{n}$
      & $\Theta(Tn\ell)$ & $\OrgState{n}{n}$ \\
    Mesh & $n(n-1)/2$ & -- & $\OrgExec{n}{n(n-1)/2}$
      & $\Theta(Tn^2\ell)$ & $\OrgState{n}{n(n-1)/2}$ \\
    Community & $m_c$ & -- & $\OrgExec{n}{m_c}$
      & $\Theta(Tm_c\ell)$ & $\OrgState{n}{m_c}$ \\
    \midrule
    G-Designer & $\widehat m_G$ & $\OrgFit{G}$
      & $\Theta(\mathcal G_G)+\OrgExec{\widehat n_G}{\widehat m_G}$
      & $\Theta(T\widehat m_G\ell)$
      & $\Theta(\mathcal M_G)+\OrgState{\widehat n_G}{\widehat m_G}$ \\
    ARG-Designer & $\widehat m_A$ & $\OrgFit{A}$
      & $\Theta(\mathcal G_A)+\OrgExec{\widehat n_A}{\widehat m_A}$
      & $\Theta(T\widehat m_A\ell)$
      & $\Theta(\mathcal M_A)+\OrgState{\widehat n_A}{\widehat m_A}$ \\
    GoAgent & $\widehat m_O$ & $\OrgFit{O}$
      & $\Theta(\OrgCall{s_g}+\mathcal G_O)+\OrgExec{\widehat n_O}{\widehat m_O}$
      & $\Theta(T\widehat m_O\ell)$
      & $\Theta(\mathcal M_O)+\OrgState{\widehat n_O}{\widehat m_O}$ \\
    \bottomrule
  \end{tabular}%
  }
  \vspace{2pt}
  
  \endgroup
\end{table*}

Table~\ref{tab:organization_complexity} adapts the supplied theoretical
complexity table. $T$ denotes exchanges rather than the role set $R$ in
Sec.~\ref{sec:notations}; $b_{\rm all}$ denotes the complete single-agent
input rather than the execution budget $B$. The call cost
$\OrgCall{s}$ includes a fixed generation allowance. Initialization, a
separate final synthesis, tool execution, and shared-state access must be
accounted for separately when material to measured costs.

Equation~\ref{eq:mas_execution} uses mean message lengths and balanced
contexts. Without this approximation, the exchange cost is expressed as
\begin{equation}
\Theta\!\left(\sum_{t=1}^{T}\sum_{i=1}^{u}
\OrgCall{b_i+\sum_{j\in\mathcal N(i)}\ell(h_j^{(t-1)})}\right).
\label{eq:mas_execution_unbalanced}
\end{equation}
Thus a nonlinear call-cost function and unequal degrees can distinguish
Star, Chain, and Tree even when their edge counts agree. The table's
balanced-context expression is not a universal hardware complexity bound.
If each edge carries messages in both directions at every exchange,
routed volume is $\Theta(Tv\ell)$. Sparse/selective routing instead sums
realized deliveries. Trace state is
$\OrgState{u}{v}=\Theta(ub+T(u+v)\ell)$ under comparable local-output
and message lengths; for a connected graph with $u\geq2$, $v\geq u-1$
reduces this to $\Theta(ub+Tv\ell)$. This includes local outputs even
when a learned graph contains isolated agents. The bound excludes backbone
parameters and KV caches and does not model measured device memory.

\textbf{Designer-generation terms.}
Let $d_h$ be hidden width, $L_G$ the GNN encoder depth, $m_0$ the template
graph's edge count, $q_r$ the role-pool size, $K_g$ the number of candidate
groups, and $g$ the number selected. Under the reference implementations
assumed by the supplied analysis, the generation costs are
\begin{align}
\mathcal G_G&=\Theta\!\left(L_G(m_0d_h+nd_h^2)+n^2d_h\right),\\
\mathcal G_A&=\Theta\!\left(nq_rd_h+n^2d_h^2\right),\\
\mathcal G_O&=\Theta\!\left(gK_gd_h+g^2d_h^2\right).
\label{eq:designer_generation_costs}
\end{align}
The G-Designer expression assumes a message-passing VGAE encoder with a
dense pairwise decoder. ARG-Designer counts $n$ role choices and
$n(n-1)/2$ autoregressive candidate-edge decisions. GoAgent counts $g$
selections over $K_g$ groups and $g(g-1)/2$ group-edge decisions; its
LLM-based group-discovery prompt of length $s_g$ is shown separately in
the online column. These are implementation-dependent reference bounds;
sparse decoders replace dense pairwise terms by evaluated candidate edges.

The corresponding auxiliary states are
\begin{align}
\mathcal M_G&=\Theta(nd_h+m_0+n^2),\\
\mathcal M_A&=\Theta(q_rd_h+nd_h+n^2),\\
\mathcal M_O&=\Theta(K_gd_h+gd_h+g^2).
\end{align}
For $N_{\rm tr}$ examples and $E_{\rm tr}$ epochs, designer fitting is
$\OrgFit{i}=\Theta(N_{\rm tr}E_{\rm tr}\mathcal G_i)$ when each example
incurs a comparable forward/backward generation cost. Backbone training,
reward evaluation, topology search, and MAS supervision rollouts are outside
this expression. Methods using different objectives or training pipelines
must report those additional costs empirically. The comparison separates
routed tokens, backbone computation, and designer work; none is a direct
substitute for measured latency or total API billing.

\subsection{Input-Perturbation Stress-Test Protocol}
\label{app:q4_data_protocol}

\textbf{Experiment scope.}
The stress test in Fig.~\ref{fig:q4_data_robustness} uses Information Integration
because complementary and conflicting inputs have explicit semantics.
The six domains and six fixed organizations are retained, with the primary
backbone, roles, allocation rules, rounds, and budgets frozen.

\textbf{Perturbations.}
For a pool of $k$ atomic input units and
$p\in\{0,0.1,0.2,0.3,0.4,0.5\}$, missing inputs remove a proportion
$p$ while retaining at least one unit. Distractors add approximately $pk$
units that resemble the task context but neither support the gold answer
nor explicitly contradict the supplied information. Conflicts introduce false versions
of selected input units through minimal numerical, entity, negation, temporal,
or causal changes, validated against the unchanged question and target.
Gold answers and support labels remain evaluator-only. Unit IDs, truth
status, and before/after text are logged.

Targeted corruption applies the same count of conflicting units after
allocation, prioritizing Star's hub, Chain's high-betweenness relay, Tree's
root/internal relays, or Community's inter-group bridge. Ring targets rotate
and Mesh targets follow a frozen seed because their vertices are structurally
equivalent. Target nodes keep executing; only their inputs change. This
is a placement stress test, so cross-topology differences combine structure
with the topology-specific targeting policy. Non-targeted perturbations
are generated before allocation and shared by all graphs for a given task,
type, ratio, and corruption seed. A single clean condition is shared across
the four perturbation types.

\textbf{Aggregation and uncertainty.}
Equation~\ref{eq:data_quality_retention} is calculated within each domain
before the six-domain macro average. The denominator must be positive;
undefined ratios are reported explicitly rather than assigned a favorable
value. Let $\overline{\mathrm{Ret}}_{o,c}(p)$ denote this macro average.
Normalized robustness AUC is
\begin{equation}
\mathrm{RAUC}_{o,c}
=\frac{1}{p_{\max}}\int_0^{p_{\max}}
\overline{\mathrm{Ret}}_{o,c}(p)\,dp,
\qquad p_{\max}=0.5,
\label{eq:data_robustness_auc}
\end{equation}
computed by trapezoidal integration on the six specified ratios and reported
on the retention-percentage scale. The displayed summary gives Mesh
$\mathrm{RAUC}=94.7$ under missing inputs and $92.3$ under targeting;
Community reaches $93.1$ with distractors and $89.5$ with conflicts.

Pointwise bands show bootstrap intervals over domain/seed records.
For paired task-level inference, resample source-task clusters with all
organizations, corruption conditions, and seeds kept together. Recompute clean
denominators and equal-domain means per replicate, then report paired
differences with 95\% intervals. Preserve incomplete and failed runs under
the shared scoring rule. Raw quality, retained quality at $p=0.5$, failure
rate, actual tokens, and message traces distinguish input loss, error
propagation, and unsuccessful conflict resolution.

\end{document}